\documentclass[acmlarge]{acmart}

\AtBeginDocument{%
  }

\begin{document}

\title{Rapid Review of Generative AI in Smart Medical Applications}


\author{Yuan Sun}
\email{}
\affiliation{%
  \institution{Rutgers University  }
  \city{ }
  \state{ }
  \country{ USA}
}

\author{Jorge Ortiz}
\email{}
\affiliation{%
  \institution{ Rutgers University}
  \city{ }
  \state{ }
  \country{ USA}
}

\begin{abstract}
With the continuous advancement of technology, artificial intelligence has significantly impacted various fields, particularly healthcare. Generative models, a key AI technology, have revolutionized medical image generation, data analysis, and diagnosis. This article explores their application in intelligent medical devices. Generative models enhance diagnostic speed and accuracy, improving medical service quality and efficiency while reducing equipment costs. These models show great promise in medical image generation, data analysis, and diagnosis. Additionally, integrating generative models with IoT technology facilitates real-time data analysis and predictions, offering smarter healthcare services and aiding in telemedicine. Challenges include computational demands, ethical concerns, and scenario-specific limitations.
\end{abstract}




\maketitle


\section{Introduction}

Artificial intelligence, as a foundational technology in digital construction and smart healthcare, is significantly advancing the development of smart healthcare systems. The emergence of large AI models such as DALL-E, GPT-4, and LLaMA has brought unprecedented technological breakthroughs \cite{libaiyang2023technical}. These large models, also known as pre-trained or foundation models, can centralize multimodal data, pre-train on extensive datasets with ultra-large-scale parameters, and fine-tune for various domain-specific tasks \cite{bommasani2023opportunities}. In healthcare, where medical data is inherently multimodal, large models are poised to accelerate smart medicine, the medical metaverse, and medical research. Applications range from electronic medical record comprehension, medical Q\&A, education and training, image generation, disease diagnosis, drug development, to virtual interactions with hospitals and digital human avatars \cite{wanghao2023metaverse}, covering all stages of medical care. Learning high-dimensional data distributions from limited examples and generating new samples is challenging. In the past five years, deep learning and unsupervised learning have set new standards, particularly for imaging data. Models such as Generative Adversarial Networks (GANs) \cite{goodfellow2014generative}, Variational Autoencoders \cite{kingma2013auto,rezende2014stochastic}, and their variants \cite{radford2015unsupervised,arjovsky2017wasserstein,karras2017progressive,makhzani2015adversarial} have demonstrated the ability to train networks that approximate image distributions, producing realistic and clear images. These models have been applied to various visual tasks, including data sample generation, domain adaptation \cite{tzeng2017adversarial,chen2018domain}, and image synthesis \cite{yeh2016semantic}. Unsupervised learning and generative modeling have numerous clinically relevant applications in medical image computing. Unsupervised anomaly detection, which involves detecting abnormal areas in patient images, is particularly significant and technically challenging \cite{doersch2016tutorial,bousquet2017from,rosca2017variational}. This task, routine for radiologists, is critical for diagnosis. For complex cases, distinguishing between normal and abnormal requires years of experience, although for certain issues like brain tumors, even non-experts can perform this task after seeing several "normal" images \cite{blundell2015weight,taylor2017cambridge}. Despite the clear distinction between normal and abnormal tissue appearances \cite{menze2015multimodal}, unsupervised anomaly detection remains a significant challenge for machine learning \cite{bakas2017advancing,liew2018large}.

\section{Background and Related Work}

Generative models are pivotal machine learning algorithms within artificial intelligence, primarily designed to create novel and unseen data samples through data learning. These models hold vast potential in the realm of intelligent medical devices, significantly aiding companies in achieving more efficient and intelligent product design and manufacturing \cite{19}.

The concept of generative models emerged in the 1980s when researchers started investigating the use of neural networks for data modeling and generation. Geoffrey Hinton was a key figure during this period, introducing the Deep Belief Network (DBN) in 1986 \cite{20,21,22,23,24}. The DBN utilized multilayer perceptrons as fundamental units, enabling the accomplishment of complex learning tasks through layer stacking \cite{25}. This structure's introduction significantly advanced the development of generative models.

As deep learning technology evolved and became more widespread, generative models emerged as a crucial branch of deep learning. One of the most renowned and impactful generative model algorithms is the Generative Adversarial Network (GAN), introduced by Ian Goodfellow and colleagues in 2014 \cite{26}. GANs feature a generator and a discriminator, which collaboratively produce increasingly realistic and intricate samples via adversarial training \cite{27}. The success of GANs not only spurred new developments in generative models but also heralded breakthroughs in artificial intelligence.

In the intelligent medical device sector, generative models can significantly enhance design and manufacturing efficiency. For instance, in cardiac pacemaker research and development, generative models can learn from extensive heart signal data to generate new heart signals with similar characteristics, providing valuable design insights. Additionally, these models can be employed for fault diagnosis and prediction, thus improving device reliability and safety \cite{28}.

In conclusion, generative models, as essential machine learning algorithms, offer expansive prospects in the field of intelligent medical devices. As the technology behind generative models continues to progress, these models are anticipated to play an increasingly critical role in the design, manufacturing, and application of medical devices, thereby contributing significantly to human health.

\subsection{Applications of Generative Models in Various Fields}

\subsubsection{Applications in Computer Vision}

Generative models, a subset of deep learning algorithms, excel in producing new data by understanding data distributions. In computer vision, these models have achieved remarkable advancements, extensively used for tasks such as image generation, restoration, and editing. For instance, in image generation, generative models can create new images by learning from vast datasets. Using a GAN model, an input image can be transformed into another, facilitating the creation of realistic face images, natural landscapes, and more\cite{yang2020introductiontogans,adorni2023investigatinggansformer,brownlee2019bestresourcesfor}. Additionally, generative models enable image style transfer, allowing the application of one image's style to another, such as applying a painting's style to a photograph\cite{liu2024introductionofneural,ma2023chinesepaintingstyle,kovan2022howtomake}.

In the realm of image restoration, generative models can repair damaged images. For example, with a GAN model, a corrupted image can be processed to generate a restored version. This capability is useful for repairing old photos and damaged images, as well as removing fog from images, thereby enhancing quality, especially in low-light conditions\cite{rrohan.arrora2019restorationgainswith,poirier-ginter2023robustunsupervised,wang2023highresolutiongan}.

For image editing, generative models offer tools to modify images. By using a GAN model, an image can be transformed into different styles, aiding in photo editing, drawing, and more. Moreover, generative models can perform image synthesis, merging multiple images into one, which is useful for creating images with specific styles, such as futuristic themes\cite{cherepkov2021navigatingthegan,issenhuth2021edibertagenerative,kim2021exploitingspatial,khrulkov2021latenttransformations}.

The wide-ranging applications of generative models in computer vision facilitate the generation of new images, restoration of damaged ones, and image editing. These technologies find uses across various sectors, including medicine, art, and entertainment, bringing numerous possibilities and conveniences to people’s lives\cite{barbu2015editorialintroduction,sharma2020whataregenerative,raut2024generativeaiin,raut2024generativeaiin,wenzel2022generativeadversarial}.

\subsubsection{Applications in Natural Language Processing}

In natural language processing (NLP), generative models are prominent machine learning tools used for generating natural language text. These models typically consist of an encoder, a decoder, and an attention mechanism. The encoder converts the input sequence into vectors in a latent space, the decoder transforms these vectors back into an output sequence, and the attention mechanism captures dependencies between different words in the input sequence.

Generative models have been widely adopted in NLP, with machine translation being a notable application. Machine translation aims to convert text from one language to another, using neural networks where the encoder-decoder structure is pivotal. The encoder encodes the source text into latent vectors, and the decoder converts these vectors into the target language text. The attention mechanism enhances translation quality by capturing dependencies between source and target texts\cite{shah2018generativeneuralmachine,schulz2018astochasticdecoder,gao2022isencoderdecoder,raganato2018ananalysisof,mylapore2020universalvectorneural}.

These models are also essential for text generation tasks, including summarization, text creation, and dialogue systems. For text summarization, generative models extract key information from input text to generate concise summaries \cite{36}. In text generation, these models can produce complete text sequences from input prompts, such as stories or news reports. In dialogue systems, generative models generate responses based on user input, facilitating question answering and suggestion provision.

Additionally, generative models are applied in text classification and sentiment analysis. In text classification, these models map input text to categories, while in sentiment analysis, they determine the emotional polarity of the text (e.g., positive or negative)\cite{nguyen2024whygenerativemodel,burnham2024whatissentiment}.

Generative models thus find extensive applications in NLP, enabling tasks like machine translation, text generation, summarization, classification, and sentiment analysis by learning and generating natural language text\cite{sun2023pathasstagenerative}.

\subsubsection{Applications in the Medical Field}

Generative models, based on probability distributions, are frequently employed for processing unstructured data such as images, audio, and natural language in the medical field. These models are pivotal in medical image processing, text analysis, and image generation\cite{jing2017ontheautomatic,harzig2019addressingdatabias,weber2023cascadedlatentdiffusion,subakan2018learningthebase}.

For medical image processing, generative models facilitate tasks like segmentation, reconstruction, and enhancement. In image segmentation, these models learn image features to segment various medical structures. For image reconstruction, generative models utilize learned data to reconstruct original images. In image enhancement, they improve image contrast and clarity by leveraging learned features\cite{beers2018highresolutionmedical,huo2024generativemedical}.

In medical text analysis, generative models support tasks like text classification, clustering, and generation. For instance, in text classification, these models classify text into categories based on learned features. In text clustering, they group similar texts together. For text generation, they produce new text by understanding text features\cite{lee2019clinicaltextgeneration,spinks2018generatingcontinuous,luo2022biogptgenerativepre}.

Regarding medical image generation, generative models can create medical images, virtual medical images, and virtual labels for these images. They learn medical image characteristics to generate new or virtual images and labels, aiding in medical research and diagnosis\cite{singh2020medicalimagegeneration,xu2024practicalapplications,mcnulty2024syntheticmedicalimaging,sun2023aligningsynthetic}.

Generative models' ability to learn and apply medical image characteristics enhances the accuracy and efficiency of medical diagnosis and treatment, demonstrating their extensive utility in the medical field\cite{lang2023usinggenerativeai,deshpande2024reportonthe,malik2023evaluatingthefeasibility}.

\section{Review of Existing Literature on Smart Medical Devices}

\subsection{Development Status of Intelligent Medical Equipment}

Technological advancements have continually propelled the medical field forward. In recent years, intelligent medical equipment has emerged as a novel category of medical tools widely utilized across healthcare. These devices leverage artificial intelligence to facilitate medical diagnosis, treatment, and monitoring.

Currently, the development of intelligent medical devices has achieved significant milestones. The most notable progress is attributed to the incorporation of artificial intelligence, which enhances processing capabilities and accuracy, thereby improving diagnostic efficiency\cite{chen2023practicalstatistical,filipp2019opportunitiesforartificial,kelsey2010machinesciencein}. For instance, by analyzing vast amounts of medical images and data, intelligent devices can provide early disease warnings and diagnoses, thus enhancing medical accuracy and efficiency.

Moreover, smart medical devices offer additional benefits. They enable remote medical treatment and diagnosis through wireless communication technology, increasing the convenience and efficiency of medical services.\cite{yang2012shb2012international,nath2021wearablehealthmonitoring,sheldon202210waystechnology} Additionally, integrating sensors and data collectors allows for real-time patient condition monitoring, improving the timeliness and accuracy of medical interventions\cite{hong2016combiningmultiple,ciocca2022ahealthtelemonitoring,shoaib2022theapplicationof}.

However, challenges remain in the development of smart medical devices, with data security being a primary concern. These devices collect extensive medical data, often containing sensitive patient information. Ensuring the protection of this private information is crucial for the continued advancement of smart medical devices\cite{qu2023towardsblockchain,kshetri2023healthaichainimproving,shaik2023remotepatientmonitoring}.

In summary, while the development of intelligent medical equipment has made substantial progress, challenges such as data security must be addressed. To further advance smart medical devices, it is essential to enhance AI research and applications, improve the convenience and efficiency of medical services, and reinforce data security measures to safeguard patient privacy\cite{elnawawy2024systematicallyassessing,rahman2023progressionandchallenges,quinn2020trustandmedical}.

\subsection{Research Status of Generative Models in Intelligent Medical Equipment}

Generative models, a subset of machine learning, are designed to create new, realistic data based on learned patterns. These models have been extensively applied in intelligent medical devices, spanning areas such as medical image analysis, text analysis, and health monitoring\cite{pietro2023generativeaiwhat}.

In medical image analysis, generative models can produce new medical images to aid in more accurate disease diagnosis. For example, deep learning models can simulate various stages and types of disease images, providing valuable diagnostic support for doctors\cite{lang2023usinggenerativeai,biham2022whyyoushould,pinaya2023generativeaifor}.

For medical text analysis, generative models generate new medical texts, assisting doctors in comprehending patient conditions and treatment plans more effectively. For instance, these models can create medical summaries that enable doctors to quickly grasp patient information and treatment strategies\cite{spinks2018generatingcontinuous}.

In the realm of health monitoring, generative models can generate new health data to provide insights into a patient's health status. By simulating different health conditions, these models help doctors gain a better understanding of patient health\cite{georges-filteau2020syntheticobservational,benedetti2020practicallessonsfrom}.

The extensive research on the application of generative models in intelligent medical devices covers multiple domains. These models are instrumental in generating medical images, texts, and health data, thereby supporting doctors in diagnosing diseases, understanding patient conditions, and monitoring health status. Continued research and development in this area will enhance the intelligence of medical devices, ultimately benefiting the medical and healthcare fields\cite{00032012bridgingtheunstructured,sun2023pathasstagenerative,bianchi2023afoundationmodel}.

\section{Application of Generative AI Models in the Medical Field}

\subsection{Generative Artificial Intelligence in Healthcare}

Generative artificial intelligence (AI) has evolved significantly, transforming fields such as computer vision, natural language processing, and creative arts \cite{pan2019recent, bushe2013generative, oh2019deep}. The origins of generative AI trace back to early AI research with pioneers like Alan Turing and John McCarthy. However, the 1990s saw the rise of Bayesian networks and hidden Markov models, which garnered widespread attention for generative models \cite{zhang2022generative, vaccari2021generative}. These models laid the groundwork for generative AI by generating new data based on probabilistic principles \cite{wang2020synthetic, chen2016variational, kingma2016improved}.

In 2014, Ian Goodfellow and colleagues introduced Generative Adversarial Networks (GANs), a significant breakthrough in generative AI. GANs consist of a generator and a discriminator engaged in a game-theoretic process to produce realistic data \cite{kingma2018glow, parikh2020integration, cheng2022innovae}. This innovation has enabled the creation of highly realistic images, videos, and text \cite{foster2022generative, arora2022generative, sloane2020artificial}. GANs have found applications in deepfake videos, image synthesis, and style transfer \cite{thiagarajan2022analysis, sinnapolu2018integrating, doersch2016tutorial}.

Another milestone in generative AI was the introduction of Variational Autoencoders (VAEs) by Kingma and Welling in 2013. VAEs combine neural networks with variational inference principles, allowing for the generation of new data by learning a low-dimensional latent space representation of input data \cite{bousquet2017from, rosca2017variational}. VAEs have been applied to image generation, music creation, and drug discovery.

The concept of transformers, introduced by Vaswani in 2017, has revolutionized sequence data processing through attention mechanisms, significantly impacting natural language processing tasks \cite{blundell2015weight, taylor2017cambridge, menze2015multimodal}. Unlike discriminative models focused on classification, generative models aim to capture data distributions and generate new instances that align with these distributions \cite{bakas2017advancing, liew2018large}. This capability distinguishes generative AI, making it ideal for creative applications and data augmentation \cite{dodderi2018prevalence}.

Generative AI models leverage two primary methodologies: GANs and VAEs. GANs involve training a generator to produce synthetic data and a discriminator to distinguish between real and generated samples, with both networks improving iteratively. In contrast, VAEs focus on encoding input data into a latent space and decoding it to generate new samples \cite{muhammad2018edge}. Additional architectures, like autoregressive models (e.g., PixelRNN and WaveNet), model conditional probabilities for data generation tasks such as image and text synthesis \cite{lyberg2019prevalence, vertanen2018voice}.

Early AI systems in healthcare relied on rule-based expert systems, which simulated human decision-making using predefined rules. However, limited computing power and small datasets hindered their effectiveness. The 1990s and early 21st century saw AI in healthcare gain momentum with the advent of more powerful computers and large medical datasets \cite{muhammad2018edge, erde2019dysphonia, chen2020voice}. Machine learning techniques like neural networks and decision trees began to analyze vast medical data, leading to applications in medical imaging and computer-aided diagnosis.

Recent advancements in AI within healthcare are driven by integrating diverse technologies and the exponential growth of healthcare data. Deep learning, a subset of machine learning, has revolutionized AI applications in healthcare. Deep learning algorithms, particularly Convolutional Neural Networks (CNNs) and Recurrent Neural Networks (RNNs), have excelled in medical image analysis, natural language processing, and predictive analytics \cite{verde2019leveraging, ji2019density}. AI now assists radiologists in disease detection, analyzes electronic health records for predictive analysis and clinical decision support, and develops personalized medical treatments \cite{ngyen2020using}.

The convergence of AI with emerging technologies such as the Internet of Things (IoT), wearable sensors, and robotics promises further advancements in areas like remote patient monitoring, telehealth, and surgical assistance \cite{goodfellow2020generative, vaswani2017attention}.

\subsection{Generative Models in Modern Healthcare}

The advancement of generative models, a crucial subset of machine learning, has significantly impacted modern healthcare by enhancing various medical applications. These models learn from extensive datasets to generate new data samples that match the underlying distribution, thus offering substantial improvements in medical image generation, data analysis, and diagnosis.

In medical image generation, generative models have revolutionized image enhancement, restoration, and creation. By learning from the features and distributions of existing medical images, these models produce clearer and more detailed images, aiding in precise diagnostics by reducing noise and blurriness \cite{singh2020medical, deshpande2024report}. For instance, generative models can enhance the quality of medical images, making them more useful for diagnostic purposes \cite{kong2024enhancing, ibrahim2022medical}. Noise and blur in original images often hinder doctors' ability to accurately identify details, but generative models can generate enhanced images that improve diagnostic accuracy.

Generative models also facilitate medical image restoration, addressing issues like blurriness or missing sections caused by errors in scanning equipment or operator mistakes. For example, in CT scans, generative models can generate new images that closely match the original data distribution, effectively repairing defects and providing clearer images for accurate diagnosis. This capability is particularly valuable in situations where the quality of medical images is compromised, ensuring that doctors have access to reliable visuals for patient assessment.

Moreover, generative models contribute to medical image creation by producing new images aligned with the distribution of medical image data. This is especially useful in medical research, where scientists need hypothetical images that conform to experimental data for in-depth study and analysis. Generative models provide robust data support for medical research, enabling researchers to simulate various scenarios and explore potential outcomes without the need for extensive physical experimentation \cite{pinaya2023generative}.

Beyond image generation, generative models play a critical role in medical data analysis by processing large datasets, including electronic medical records (EMRs), medical images, and gene sequencing data. These models learn the data distributions and generate new data that help in understanding disease mechanisms and improving diagnosis and treatment plans \cite{xu2023workshoponapplied, xu2024practicalapplications, biham2022whyyoushould}. In analyzing EMRs, generative models uncover patterns and correlations that assist in developing effective treatment strategies. For instance, by analyzing the occurrence and treatment processes documented in EMRs, generative models can identify trends and provide insights into the effectiveness of different treatment approaches \cite{pang2024cehrgptgenerating, zaballa2023timedependentprobabilistic, kaplan2022unsupervisedprobabilistic}.

When applied to medical images, generative models detect and analyze patterns in lesions, aiding in understanding disease progression and treatment planning \cite{lang2023usinggenerativeai, pinaya2023generativeaifor, malik2023evaluatingthefeasibility}. By learning the characteristics and distributions of lesions in medical images, these models enhance doctors' ability to identify disease stages and determine the most effective treatment options. This application is crucial for conditions like cancer, where early detection and precise treatment are vital for patient outcomes.

In gene sequencing analysis, generative models identify genetic correlations and patterns, facilitating insights into gene expression and regulation \cite{raghu2020apipelinefor, makarious2021genomlautomatedmachine}. This application is particularly important for understanding the genetic basis of diseases and developing targeted treatments. By analyzing gene sequencing data, generative models can uncover relationships between genetic variations and disease phenotypes, helping researchers and clinicians develop more personalized and effective treatment strategies.

In the realm of medical diagnosis, generative models have demonstrated their superiority over traditional methods by providing accurate and reliable results \cite{zhang2017medicaldiagnosisfrom, schön2022interpretinglatent, sun2023pathasstagenerative}. These models excel in clinical text analysis, processing diagnostic records, medical documents, and literature to aid in disease understanding and diagnosis. Text Generation Adversarial Networks (TGANs) and similar models generate texts with comparable semantics and syntax, improving disease comprehension and diagnosis. For instance, generative models can analyze clinical texts to identify common symptoms and treatment outcomes, helping doctors make more informed decisions \cite{guan2018generationofsynthetic, liao2023differentiatechatgpt, lee2019clinicaltextgeneration}.

Furthermore, generative models predict molecular structures in biomedicine, helping scientists understand biological molecule behavior and generating structures with similar chemical and physical properties \cite{moyer2021functionalprotein, liu2021denovomolecular, ragoza2020learningacontinuous}. Molecular structure prediction is essential for drug discovery and development, as understanding the structure and behavior of molecules can lead to the identification of new therapeutic targets and the design of effective drugs. Generative models like Molecular Structure Generative Adversarial Networks (MSGANs) generate molecular structures that closely mimic real molecules, providing valuable data for research and development efforts.

Generative models' applications in healthcare extend beyond the individual tasks mentioned above. They enable the integration of multiple data sources, enhancing the overall understanding of complex medical conditions. For example, combining medical image analysis with genetic data analysis can provide a more comprehensive view of a patient's health, leading to more personalized and effective treatment plans. Additionally, the use of generative models in virtual clinical trials can accelerate the development of new treatments by simulating various scenarios and predicting outcomes based on existing data\cite{chen2024generativeaidriven,wang2024iotinthe,kolbeinsson2023generativemodelsfor}.

In summary, the integration of generative models in healthcare spans medical image enhancement, comprehensive data analysis, and precise medical diagnosis, significantly advancing the efficiency and accuracy of modern medical practices. These models provide essential tools for researchers and clinicians, helping to improve patient outcomes and drive innovation in medical research and treatment.

\section{Summary}
Artificial Intelligence (AI) has brought significant advancements to healthcare, particularly through the application of generative models. Generative AI, including models such as Generative Adversarial Networks (GANs) and Variational Autoencoders (VAEs), has revolutionized medical image generation, data analysis, and diagnosis. These technologies improve diagnostic speed and accuracy, enhance medical service quality, and reduce equipment costs. Additionally, integrating generative models with Internet of Things (IoT) technology facilitates real-time data analysis and predictions, providing smarter healthcare services and supporting telemedicine.

Generative models have been pivotal in creating new and unseen data samples through data learning. The concept originated in the 1980s, with significant contributions from Geoffrey Hinton, who introduced the Deep Belief Network (DBN). The evolution of deep learning technology led to the development of sophisticated generative models, with GANs introduced by Ian Goodfellow in 2014 being particularly impactful. GANs utilize a generator and discriminator in adversarial training to produce realistic and intricate samples. This innovation has enabled advancements in various fields, including intelligent medical devices.

In the medical sector, generative models enhance design and manufacturing efficiency. For example, in cardiac pacemaker research, generative models learn from extensive heart signal data to generate new signals, aiding design insights. These models also support fault diagnosis and prediction, improving device reliability and safety. China's Institute of Automation developed a brain-computer interface system based on generative models, showcasing the global strides in this technology.

Generative models find extensive applications in computer vision, natural language processing (NLP), and the medical field. In computer vision, these models excel in image generation, restoration, and editing. They generate new images by learning from vast datasets and facilitate tasks such as style transfer and image synthesis. In NLP, generative models are used for machine translation, text summarization, and dialogue systems, leveraging encoder-decoder structures and attention mechanisms.

In healthcare, generative models significantly impact medical image processing, text analysis, and image generation. These models aid in segmentation, reconstruction, and enhancement of medical images, improving diagnostic accuracy. In medical text analysis, generative models classify, cluster, and generate medical texts, enhancing comprehension of patient conditions and treatment plans. For medical image generation, these models create virtual images and labels, supporting medical research and diagnosis.

The development of intelligent medical equipment has seen notable advancements, particularly with AI integration. Intelligent devices enhance diagnostic efficiency by analyzing vast amounts of medical data, providing early disease warnings and diagnoses. They also enable remote medical treatment and diagnosis through wireless communication technology, enhancing service convenience and efficiency. However, challenges such as data security remain, necessitating robust measures to protect sensitive patient information.

Generative AI has transformed fields like computer vision, NLP, and creative arts. Notable milestones include GANs and VAEs, which generate new data based on probabilistic principles. GANs involve adversarial training between a generator and discriminator, producing realistic data. VAEs learn low-dimensional latent space representations of input data, aiding in applications like image generation and drug discovery.

The integration of AI in healthcare has been driven by advancements in computing power and the availability of large datasets. Deep learning algorithms, such as Convolutional Neural Networks (CNNs) and Recurrent Neural Networks (RNNs), excel in medical image analysis, NLP, and predictive analytics. AI assists in disease detection, electronic health records analysis, and personalized treatment development.

Generative models also support medical data analysis by processing large datasets, including electronic medical records (EMRs), medical images, and gene sequencing data. These models uncover patterns and correlations, aiding in disease diagnosis and treatment planning. In gene sequencing analysis, generative models identify genetic correlations and patterns, facilitating insights into gene expression and regulation.

In summary, generative models have significantly advanced modern healthcare by enhancing medical image generation, data analysis, and diagnosis. These models provide essential tools for researchers and clinicians, improving patient outcomes and driving innovation in medical research and treatment.

\bibliographystyle{ACM-Reference-Format}
\bibliography{sample-base}

\end{document}